\pdfoutput=1
\documentclass[11pt]{article}

\usepackage[preprint]{acl}
\usepackage{times}
\usepackage{latexsym}
\usepackage[T1]{fontenc}
\usepackage[utf8]{inputenc}
\usepackage{microtype}
\usepackage{inconsolata}

\usepackage{amsmath,amssymb,amsfonts}
\usepackage{graphicx}
\usepackage{textcomp}
\usepackage{xcolor}
\usepackage{multirow}
\usepackage{algorithm}
\usepackage{algorithmicx}
\usepackage{algpseudocode}
\usepackage{multicol}
\usepackage{tabularx}
\usepackage{diagbox}

\title{Improving Non-autoregressive Machine Translation with Error Exposure and Consistency Regularization}

\author{Xinran Chen, Sufeng Duan, Gongshen Liu \\
School of Electronic Information and Electrical Engineering, \\Shanghai Jiao Tong University\\
\texttt{\{jasminechen123,1140339019dsf,lgshen\}@sjtu.edu.cn}\\
}
\begin{document}
\maketitle
\begin{abstract}
Being one of the IR-NAT (Iterative-refinemennt-based NAT) frameworks, the Conditional Masked Language Model (CMLM) adopts the mask-predict paradigm to re-predict the masked low-confidence tokens. 
However, CMLM suffers from the data distribution discrepancy between training and inference, where the observed tokens are generated differently in the two cases. 
In this paper, we address this problem with the training approaches of error exposure and consistency regularization (EECR). 
We construct the mixed sequences based on model prediction during training, and propose to optimize over the masked tokens under imperfect observation conditions. 
We also design a consistency learning method to constrain the data distribution for the masked tokens under different observing situations to narrow down the gap between training and inference. 
The experiments on five translation benchmarks obtains an average improvement of 0.68 and 0.40 BLEU scores compared to the base models, respectively, 
and our CMLMC-EECR achieves the best performance with a comparable translation quality with the Transformer. The experiments results demonstrate the effectiveness of our method.
\end{abstract}

\section{Introduction}

Although non-autoregressive machine translation (NAT)~\cite{gu2018non, kaiser2018fast} generates the target sentence in parallel and accelerates generation, NAT models cannot model the multi-modal distribution of the target sentence as autoregressive translation (AT)~\cite{vaswani2017attention} and hurts the translation performance~\cite{gu2018non, ghazvininejad-etal-2019-mask}. 
To better capture the multi-modality distribution of the target sentence, 
\cite{lee2018deterministic} and \cite{ghazvininejad-etal-2019-mask} proposed iterative-refinement-based NAT (IR-NAT) models, which refine the target sentence through multiple rounds of prediction. 
Especially, Conditional Masked Language Model-based (CMLM) NAT ~\cite{ghazvininejad-etal-2019-mask,ghazvininejad2020semi,kasai2020non,guo2020jointly} is one kind of IR-NAT. 
CMLM adopts a mask-predict decoding strategy to capture the interdependency within the target sentence, in which CMLM masks the low-confidence tokens of the previous round and re-predicts them in the next round. 

However, CMLM still suffers from the data distribution mismatch between training and inference~\cite{ghazvininejad2020semi,huang2021improving}. 
In detail, the masked tokens in the decoder input are predicted based on the observed tokens in both training and inference, while the observed tokens are from ground truth and model predictions in training and inference, respectively. 
This data distribution discrepancy, called \textit{exposure bias}~\cite{ranzato2015sequence}, could impair the model performance. 
To solve the problem, some methods have been proposed. 
One line of work utilizes consistency learning to diminish the output mismatch in different situation.  
MvSR-NAT~\cite{xie2022mvsr} designs multi-view subset regularization for the CMLM model which makes NAT models consistent at the level of shared masks and model parameters. 
However, the observed words in MvSR-NAT are all ground truth tokens, and it neglects the potential errors in inference. 
Another line of work exposes the model to inference errors during training. 
For example, CMLMC~\cite{huang2021improving} and CMLM-SMART~\cite{ghazvininejad2020semi} both propose to introduce potential errors and correct them within the observed tokens. 
However, the mismatch still exists. 
For CMLMC, it ignores to update the masked tokens in the erroneous observing range, which is necessary in inference. 
And for CMLM-SMART, the potential error distribution during training is not consistent with that under inference, as the potential errors in observing range during training are generated based on the masked ground truth, and the errors in inference are produced based on the previous predicted sequence.

To address the problem, we propose the approaches of \textbf{E}rror \textbf{E}xposure and \textbf{C}onsistency \textbf{R}egularization (EECR) to shrink the data distribution mismatch between training and inference of CMLM in this paper. 
First, we propose a method to supervise the model training with error exposure, in which we replace a portion of the observed tokens of the ground truth with the predicted one to construct a mixed sequence, and then optimize the model over every masked token. 
Since the predicted tokens in training examples are generated by multi-step refinement, the data distribution of training gets closer to that of inference, thus alleviating the data exposure bias. 
Second, we set consistency regularization as an auxiliary optimization objective during training, which requires the probability distribution for the masked tokens to be consistent under different scenarios. 
This approach not only enhances the consistency of the prediction distribution between training and inference, but also improves the model robustness.

We apply EECR to CMLM and CMLMC, and validate our model on five datasets, WMT14 EN$\leftrightarrow$DE, WMT16 EN$\leftrightarrow$RO and IWSLT14 DE$\rightarrow$EN. 
Experimental results demonstrate the generality of our models with an average improvement of 0.68 and 0.40 BLEU scores on distilled datasets compared to the baselines. 
Especially, our model outperforms several strong NAT competitors in terms of translation quality and CMLMC-EECR obtains comparable performance. 

The contributions of this paper are summarized as follows: 

(i) Our method alleviates the mismatch between training and inference for mask-predict-based NAT models by introducing potential inference errors and adding consistency regularization during training. 

% The method constructs training examples containing errors and optimizes the model under imperfect conditions. 
% In addition, we adopt consistency regularization to make the probability distributions of the masked tokens in different observing conditions as consistent as possible,

(ii) Our method has great generality and could be applied to different kinds of conditional masked language models. 

(iii) The experimental results in five datasets reveal that our EECR strategy could improve the translation quality for the base model and further shrink the performance gap with the AT model.

\section{Related Work}
Since the NAT decoding paradigm was proposed, various methods have been put forward to enhance the quality of non-autoregressive translation. 
Existing NAT models can be categorized into fully NAT models and iterative-refinement-based NAT models according to their decoding patterns.
Fully NAT models produce translation in a single round. 
Vanilla-NAT~\cite{gu2018non} model is the first work of fully NAT, but it fails to capture the target-side dependency. 
To overcome this problem, various approaches have been proposed, such as 
latent variables-based NAT models ~\cite{shu2020latent, zhu2022non, ma2019flowseq, bao2021non}, improved cross-entropy as loss function~\cite{ghazvininejad2020aligned, du2021order, li-etal-2022-multi-granularity, shao2020minimizing, du2022ngram}, 
target-side dependency-based models~\cite{sun2019fast, ran2021guiding, song2021alignart} and enhanced learning strategy-based models~\cite{qian2021glancing, huang2022non, zhan2023depa, guo2023renewnat}.
Iterative-refinement-based NAT models refine the translation through multiple rounds of iterations to better capture the dependencies within the target sentence~\cite{lee2018deterministic}. 
Iterative-refinement-based NAT models include insertion-deletion-NAT~\cite{stern2019insertion, gu2019levenshtein}, CMLM~\cite{ghazvininejad-etal-2019-mask}, etc. 
There are quite a few methods to improve the CMLM, including additionally masking the source-side sequences to enhance the modeling of source-side embedding~\cite{guo2020jointly,xiao2023amom}, designing self-correction mechanism by introducing an auxiliary decoder to judge the correctness of the NAT  outputs~\cite{xie2020infusing,geng2021learning}, adjusting the visible range of masked tokens to make the contexts more diversified~\cite{kasai2020non}, as well as introducing advanced learning strategies including multi-task learning~\cite{hao2021multi}, contrastive learning~\cite{cheng2022nat}, and so on.

Similar to previous methods~\cite{ghazvininejad2020semi,huang2021improving}, we expose the model to prediction errors during training. 
However, for the supervised objective, we predict the masked tokens under various situations, while CMLMC~\cite{huang2021improving} focuses on self-correcting the errors in observed tokens and CMLM-SMART~\cite{ghazvininejad2020semi} concentrates on supervising the prediction of the whole mixed sequence.
In addition, CMLMC and CMLM-SMART only adopt negative log likelihood loss as the optimization objective, while we further propose to utilize consistency regularization to enhance the consistency of the output distribution.

Consistency regularization is a semi-supervised method that prevents the model from overfitting as well as improves model robustness~\cite{sajjadi2016regularization, NEURIPS2021_5a66b920, laine2016temporal, NEURIPS2020_44feb009}, which has also gained applications in the field of NAT. 
CR-LaNMT~\cite{zhu2022non} utilizes consistency regularization to improve the variational autoencoder-based NAT model by injecting noise into the input and implementing consistency learning on the posterior latent variables. 
MvSR-NAT~\cite{xie2022mvsr} proposes multi-view subset regularization for the CMLM model, which argues that NAT models should be consistent at the level of shared masks and model parameters. However, the observed words in the MvSR-NAT are all ground truth tokens, and it ignores the potential errors from inference. 
Unlike CR-LaNMT for the VAE models, we improve CMLM models and augment the training data by mixing the ground truth with predictions in the observed parts as the decoder input. 
Different from MvSR-NAT, our approach exposes the model to prediction errors during training, thus diminishing the discrepancy in training and inference. 

\section{Approach}
In this section, we first introduce our EECR in detail, including the method of training with error exposure and consistency regularization. Then we present the process of model training and inference.

\begin{figure*}[htbp]
\centerline{\includegraphics[width=0.95\textwidth]{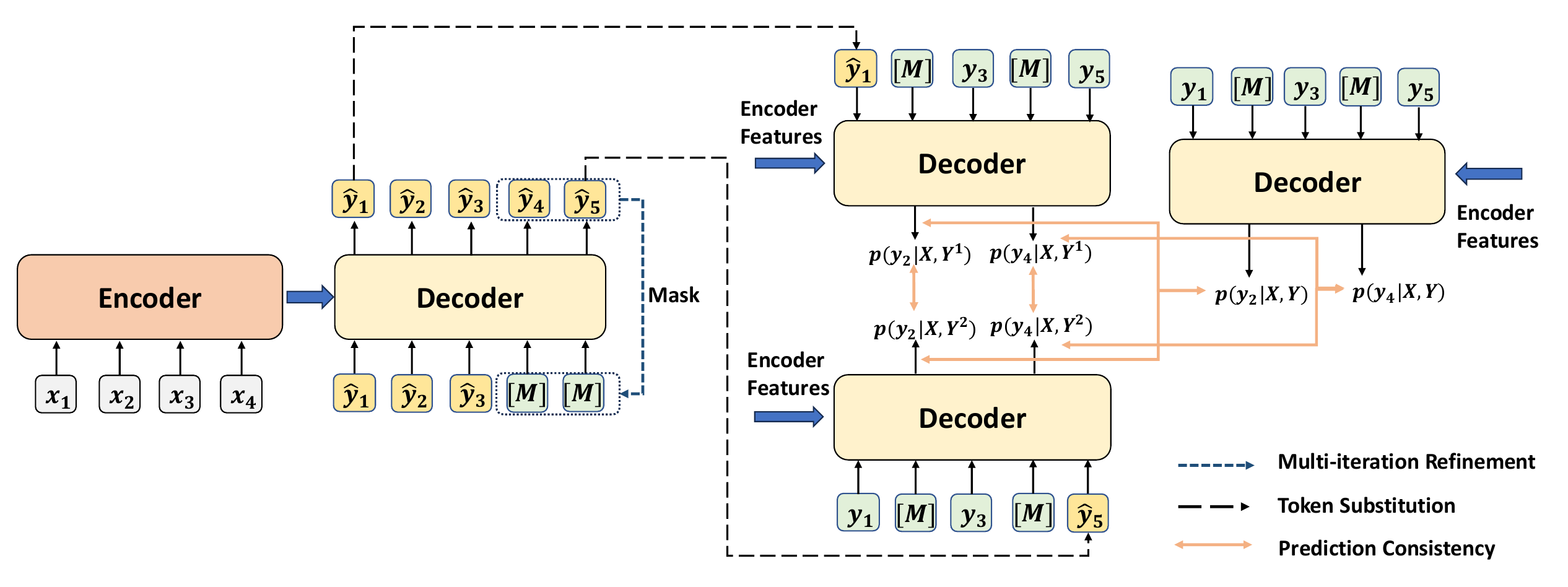}}
\caption{The overview of the our EECR strategy. 
The left part illustrates the sequence prediction process of the mixed sequence generation. 
The decoder refines the predicted sequence $\hat{Y}$ based on the sequence of the former step $\hat{Y}_{Prev}$ (as shown by the blue dotted line arrows) by $k$ times. 
Subsequently, the partially masked ground truth sequences are randomly substituted with the predicted tokens $\hat{y}_1$ and $\hat{y}_5$ (as shown by the dashed arrows) and we get the mixed sequences $Y^{1}$ and $Y^{2}$. 
The right part depicts the consistency learning process. 
The probability distributions of the masked tokens $\texttt{[M]}$ under the ground truth and mixed sequences are constrained by the consistency regularization (as shown by the bidirectional arrows).}
\label{fig1}
\end{figure*}
\subsection{Training With Error Exposure}
To narrow the data distribution gap between training and inference, the potential inference errors should be exposed to the model as much as possible during training. 
Therefore, we first construct the mixed sequence by substituting a part of the observed ground truth tokens with the predicted tokens in the decoder input. 
Based on the mixed sequence, we optimize our model with the cross-entropy loss over every masked token during training.

The construction of the mixed sequence has two steps. 
The first step is sequence prediction, in which we get the final predicted sequence by introducing the multi-step refinement. 
Specifically, we set the maximum iteration number $K$ and randomly select an iteration number $k$ from $1\sim K$ in each update to produce the predicted sequence. 
Introducing multi-step refinement instead of single-step refinement in training could expose the model to potential inference errors from different iterations and therefore better shrink the data distribution bias between training and inference. 

With the predicted sequence, the second step is token substitution. 
We denote the partially masked ground truth as $Y$ and $y_{t}$ is the t-th token in $Y$. We represent the predicted sequence as $\hat{Y}$ and $\hat{y_{t}}$ is the t-th token in $\hat{Y}$. 
We replace the $y_{t}$ among $Y$ with the predicted token $\hat{y_{t}}$ with probability $\beta$ twice, and we get two different mixed sequences, $Y^{1}$ and $Y^{2}$. 
The random substitution to get the mixed sequence $Y^{i}, i \in \{1,2\}$ could be defined as:
\begin{eqnarray}
    s&\sim&\mathrm{Uniform}(0,1),\label{eqn:uniform}\\
    y_t^{i}&=&
        \begin{cases}
              {y_t} & s > \beta\\
              \hat{y_{t}} & s \leq \beta
        \end{cases},
\end{eqnarray}
where $y_t^{i}$ represents the t-th token in the mixed sequence $Y^{i}$. 
Note that the masked positions in the ground truth are excluded from the substitution operation, and only tokens within the observing range could be substituted. The mixed sequence construction is depicted in the left part of Figure~\ref{fig1}. 

Based on the constructed mixed sequences, we supervise over the masked tokens ($Y_{mask}$) during training. The prediction of $Y_{mask}$ under potential inference errors makes the model robust to errors and shrinks the data distribution discrepancy between training and inference.

We select the negative log-likelihood (NLL) training loss functions for two mixed sequences, which are expressed as: 
\begin{equation}
\label{eqn:nll1}
\mathcal{L}_{nll}^{1} = -\sum^{\lvert Y_{mask}\rvert}_{t=1}{ \log P\left(y_1|X, Y_{pred}^{1}, Y_{obs}\backslash Y_{pred}^{1}\right)}
\end{equation}
\begin{equation}
\label{eqn:nll2}
\mathcal{L}_{nll}^{2} = -\sum^{\lvert Y_{mask}\rvert}_{t=1}{ \log P\left(y_2|X, Y_{pred}^{2}, Y_{obs}\backslash Y_{pred}^{2}\right)}
\end{equation}
where $Y_{mask}$ is the set of common masked tokens, $Y_{pred}^{i}$ and $Y_{obs} \backslash Y_{pred}^{i}, i \in \{1,2\}$ denote the set of the predicted tokens and the ground truth tokens within the observed tokens in the mixed sequence $Y^{i}, i \in \{1,2\}$, respectively.

We also keep the original CMLM loss with the observed tokens all from the ground truth, which is:
\begin{equation}
\label{eqn:nll3}
\mathcal{L}_{nll}^{3} = -\sum^{\lvert Y_{mask}\rvert}_{t=1}{ \log P\left(y_3|X,Y_{obs} \right)}
\end{equation}
where the masked token set $Y_{mask}$ is the same in Equation \ref{eqn:nll1}, \ref{eqn:nll2} and \ref{eqn:nll3}.

\subsection{Consistency Regularization}
The consistency regularization is illustrated in the right part of Figure~\ref{fig1}. 
To constrain the output distribution of the model, we introduce the symmetric KL divergence (KLD) as the optimization objective in training. 

The first KLD is the symmetric KL divergence between the probability of the masked tokens under different views of the mixed sequence: 
\begin{equation}
\label{eqn:kld1}
\begin{split}
\mathcal{L}_{kld}^{1} = -\frac{1}{n}
\sum^{\lvert Y_{mask}\rvert}_{t=1}
\frac{1}{2}\big[
\mathcal{D}_{KL}\big( P\left(y_1\right) || P\left(y_2\right) \big) + \\
\mathcal{D}_{KL}\big( P\left(y_2\right) || P\left(y_1\right) \big)
\big]
\end{split}
\end{equation}
where $y_1$ and $y_2$ represent the masked token under the mixed sequence $Y^1$ and $Y^2$, respectively. $n$ denotes the number of the masked tokens.

To make the probability distribution of the masked tokens under prediction errors consistent with that under ground truth, we introduce another KL divergence as the optimization objective during training by:
\begin{equation}
\label{eqn:kld2}
\begin{split}
\mathcal{L}_{kld}^{2} =  -\frac{1}{n} 
\sum^{2}_{i=1}\sum^{\lvert Y_{mask}\rvert}_{t=1}
\frac{1}{2}\big[
\mathcal{D}_{KL}\big( P\left(y_i\right) || P\left(y_3\right) \big) +\\ 
\mathcal{D}_{KL}\big( P\left(y_3\right) || P\left(y_i\right)\big)
\big] 
\end{split}
\end{equation}

where $y_3$ denotes the masked tokens under the ground truth observed tokens. 
With consistency regularization of Equation~\ref{eqn:kld1} and~\ref{eqn:kld2}, the model is encouraged to be robust to various errors and the probability distribution under an erroneous setting could get closer to that of ground truth.
% shrinking the discrepancy between training and inference.

\subsection{Training and Inference}
\subsubsection{Length prediction}
Unlike the AT model which uses a special token $\texttt{[EOS]}$ as a decoding terminator, the CMLM model generates translations in parallel and determines the output length before decoding. For length prediction, a length token $\texttt{[LENGTH]}$ is added to the input of the encoder, and the corresponding output is used to predict the target sentence length by a length predictor~\cite{ghazvininejad-etal-2019-mask}. The loss function for length prediction is written as:

\begin{equation}
\label{eqn:lenth}
\mathcal{L}_{len} = -\log P\left(l_Y|X,\theta\right)
\end{equation}
where $l_y$ is the length of ground truth $Y$ .

\subsubsection{Training Algorithm}
We combine the NLL losses, the KL divergence losses, and the length loss mentioned above to obtain the total loss:
\begin{equation}
\label{eqn:total}
\mathcal{L}_{total} = 
\frac{1}{3}(
\mathcal{L}_{nll}^{1}+
\mathcal{L}_{nll}^{2}+
\mathcal{L}_{nll}^{3}) 
+\frac{\gamma}{3}(
\mathcal{L}_{kld}^{1}+
\mathcal{L}_{kld}^{2}
)
+ \mathcal{L}_{len}
\end{equation}
where $\gamma$ is a hyperparameter controlling the intensity of consistency regularization. We display our training process in Algorithm~\ref{alg:transform}.

\renewcommand{\algorithmicrequire}{\textbf{Input:}}
\renewcommand{\algorithmicensure}{\textbf{Output:}}
\begin{algorithm}[htb]
  \caption{Training Algorithm} % 名称
  \label{alg:transform}
  \begin{algorithmic}[1]
    \Require
    Training pairs $\{(X,Y)\}$
    \Ensure
       Model Parameter $\theta$
\State Initialize Model Parameter  $\theta$
\While{model not coverage}
\For{($X,Y$) in Training set}
 \State  Sample $Y_{mask}$, $Y_{obs}$ from Y
 \State Predict
 $\hat{Y} = Decoder(X,\hat{Y}_{Prev})$
 \State Replace tokens in $Y_{obs}$ with $\hat{Y}$ 
 \State Get mixed sequence $Y^{1}$ and $Y^{2}$ 
 \State Compute NLL loss in Eq.~\ref{eqn:nll1}, \ref{eqn:nll2}, \ref{eqn:nll3}
  \State Compute KLD loss in Eq.~\ref{eqn:kld1} and \ref{eqn:kld2}
   \State Compute length loss in Eq.~\ref{eqn:lenth}
  \State Update $\theta$ by minimizing loss in Eq.~\ref{eqn:total}
\EndFor
\EndWhile
\end{algorithmic}
\end{algorithm}

\subsubsection{Inference}
The inference of our model is consistent with CMLM~\cite{ghazvininejad-etal-2019-mask}. 
For the first iteration, the decoder generates predictions based on a fully masked sequence $Y_{\emptyset}$ as input. 
For subsequent iterations, the decoder replaces tokens with the lowest probability in the previous round with $\texttt{[MASK]}$ as input and subsequently makes a prediction. 
The iteration continues until the maximum number of iterations is reached or the prediction is no longer updated.

\begin{table*}[th!]
\centering
\renewcommand{\arraystretch}{1.01}
\small
\caption{Performance comparison between our model and previous NAT models. \textbf{Iter.} is the number of decoding iterations, Adv. denotes adaptive decoding. $^*$ represents the results under our implementation with the distilled datasets, and the original result is displayed in Table~\ref{tab:basecompare}. \textbf{-} indicates that the data is not reported in the original paper.}
\scalebox{0.87}{
\begin{tabular}{ll|c|c|cc|cc}
\hline
\multicolumn{2}{l|}{\multirow{2}{*}{\textbf{Models}}} &
\multirow{2}{*}{\textbf{Iter.}} &
\multirow{2}{*}{\textbf{Speedup}}&
\multicolumn{2}{c|}{\textbf{WMT'14}} & \multicolumn{2}{c}{\textbf{WMT'16}} \\
\multicolumn{1}{c}{} & & & & \textbf{EN$\rightarrow$DE} & \textbf{DE$\rightarrow$EN} & \textbf{EN$\rightarrow$RO} & \textbf{RO$\rightarrow$EN} \\
\hline
\multirow{2}{*}{AT}
& Transformer (\textit{base})~\cite{vaswani2017attention}& N & 1.0$\times$ &  27.30 & 31.29 &- & -\\
& Transformer$^*$ (\textit{base}) & N & 1.0$\times$ &  \bf{28.19}&\bf{31.74}&\bf{34.14}&\bf{34.37} \\
\hline
\multirow{16}{*}{Fully NAT}
& Vanilla NAT~\cite{gu2018non} &  1 & 15.6$\times$ & 17.69 & 21.47 & 27.29 & 29.06 \\
% &BoN-Joint+FT~\cite{shao2020minimizing} &  1 & 10.77$\times$ & 20.90 & 24.61 & 28.31 &29.29 \\
&LaNMT~\cite{shu2020latent} &  1 &22.2$\times$ & 22.20 &26.76 & 29.21 & 28.89 \\
& DCRF~\cite{sun2019fast} & 1 & 10.4$\times$ & 23.44 & 27.22 &  - & -\\
& Flowseq~\cite{ma2019flowseq} & 1 & 1.1$\times$ & 23.72 &  28.39 & 29.73  & 30.72\\
& ReorderNAT~\cite{ran2021guiding} & 1 & 16.1$\times$ & 22.79 & 27.28 & 29.30 & 29.50 \\
& AXE~\cite{ghazvininejad2020aligned} & 1 & 15.3$\times$ & 23.53 & 27.90 & 30.75 & 31.54 \\
&CNAT~\cite{bao2021non} &1 & 10.37$\times$ & 25.56 & 29.36 & - & - \\
&CR-LaNMT~\cite{zhu2022non} & 1 & 21.1$\times$ & 25.59 &30.11 &31.40 &31.63 \\
& GLAT+DSLP~\cite{huang2022non} & 1 & 14.9$\times$ & 25.69 & 29.90 & 32.36 & 33.06 \\
& OAXE~\cite{du2021order} &1&15.3$\times$&26.10&30.20&32.40&33.30\\
& MgMO~\cite{li-etal-2022-multi-granularity} & 1 & 15.3$\times$ & 26.40 &30.30 &32.90& 33.60 \\
& AlignNART~\cite{song2021alignart} & 1 & 13.4$\times$  &26.40 & 30.40 & 32.50 & 33.10 \\
& ngram-OAXE~\cite{du2022ngram} &1&15.3$\times$ &26.50 &30.50 &-&-\\
& GLAT+NPD~\cite{qian2021glancing} & 1 & 7.9$\times$ & 26.55 & 31.02 & 32.87 & 33.51 \\
&GLAT+RenewNAT~\cite{guo2023renewnat}&1 & 11.2$\times$ & 26.65 &30.65 &33.02& 33.74\\
&CMLMC+DiMS~\cite{norouzi2023dims} &1 &-& 26.7 &31.1 &33.2& 33.6 \\
 &CTC+VAE~\cite{gu2021fully} & 1 & 16.5$\times$  &27.49 & 31.10& 33.79 & 33.87 \\
& CTC+DePA~\cite{zhan2023depa} & 1 &  14.7$\times$ &27.51 &31.96 &34.48& 34.77 \\
\hline
\multirow{9}{*}{Iterative NAT}
% & NAR-IR~\cite{lee2018deterministic} &10 & 1.5$\times$& 21.61 &25.48 &29.32 & 30.19 \\
% &CMLM~\cite{ghazvininejad-etal-2019-mask} & 10 & 3.77$\times$ & 27.03 & 30.53 & 33.08 & 33.31 \\
&LevT~\cite{gu2019levenshtein} & Adv. & 4.0$\times$ & 27.27 & - & - & 33.26 \\
&DisCO~\cite{kasai2020non} & Adv. & 3.5$\times$ & 
27.34 & 31.31 & 33.22 & 33.25  \\
& InsT~\cite{stern2019insertion} & $\approx$log N & 4.8$\times$ & 27.41 & - & - & - \\
% &AMOM~\cite{xiao2023amom} & 10&2.3$\times$ & 27.57 & 31.67 &34.62 & \textbf{34.82} \\
&CMLM-SMART~\cite{ghazvininejad2020semi} & 10 & 1.7$\times$ & 27.65 & 31.27 & -&- \\
&JM-NAT~\cite{guo2020jointly} & 10 & 5.7$\times$ & 27.69 & \bf{32.24} & 33.52 & 33.72\\
& RewriteNAT~\cite{geng2021learning} &Adv.&-&27.83& 31.52& 33.63 &34.09 \\
&Con-CMLM~\cite{cheng2022nat} & 10&1.7$\times$ & 27.93 &31.57& 33.88 &34.18 \\
& MvCR-NAT~\cite{xie2022mvsr} &10& 3.77$\times$ &27.94 &31.68& 33.38& 33.92 \\
& ReviewNAT~\cite{xie2020infusing} & 10 & 1.7$\times$ & 27.97 &31.59 &33.98 &\textbf{34.34} \\
&Multi-Task NAT~\cite{hao2021multi}&10&1.7$\times$&27.98&31.27&33.80&33.60\\
% &CMLMC~\cite{huang2021improving} & 10 & 1.7$\times$ & 28.37 & 31.41 & 34.57 & 34.13 \\
\cline{1-8}
\multirow{4}{*}{Ours}
&\textbf{CMLM$^*$}~\cite{ghazvininejad-etal-2019-mask} & 10 & 3.77$\times$ & 27.25 & 31.08 & 33.15 & 33.41 \\
& \textbf{CMLM+EECR} &  10 & 3.77$\times$ &27.95& 31.59& 33.84 &34.21 \\
\cline{2-8}
&\textbf{CMLMC$^*$}~\cite{huang2021improving} & 10 & 3.77$\times$ & 27.38 & 31.16 & 34.08 & 34.15 \\
& \textbf{CMLMC+EECR} &  10 & 3.77$\times$ &\textbf{28.04}& 31.65& \textbf{34.33} &34.32 \\
\hline
\end{tabular}
}
\label{tab:main}
\end{table*}

\section{Experiments}
\subsection{Setup}
\subsubsection{Dataset}
We adopt five datasets, WMT14 EN$\leftrightarrow$DE (about 4.5M), WMT16 EN$\leftrightarrow$RO (about 610k) and IWSLT14 DE$\rightarrow$EN (about 150k) to evaluate our model. 
We utilize the same train, valid, and test sets as previous works~\cite{ghazvininejad-etal-2019-mask, kasai2020non, xie2022mvsr} for a fair comparison.
For WMT14 EN$\leftrightarrow$DE and WMT16 EN$\leftrightarrow$RO, we use both the raw and distilled dataset from~\cite{ghazvininejad-etal-2019-mask, kasai2020non}, while for IWSLT14 DE$\rightarrow$EN, we utilize the original raw dataset as~\cite{xie2022mvsr}. 
Following the previous works~\cite{ghazvininejad-etal-2019-mask, kasai2020non}, we adopt BPE~\cite{sennrich2016neural} to generate shared vocabularies that consist of about 32k subwords.

\subsubsection{Sequence-level Knowledge Distillation}
Same as the previous works~\cite{gu2018non,ghazvininejad-etal-2019-mask,kaiser2018fast}, we employ the original source data with the distilled target data generated by the AT teacher model to train the NAT model. 
The sequence-level knowledge distillation reduces the modalities in the training data, which lowers the difficulty of training the NAT model.

\subsubsection{Details}
We adopt Transformer as the framework of the model and the details of the hyper-parameters are given in Appendix~\ref{app:Details}.
Regarding the substitution probability $\beta$, we set it to 0.3 based on the result of the grid search within \{0.1, 0.2, 0.3, 0.5\}.

\subsubsection{Evaluation}
For translation quality, we use BLEU~\cite{papineni2002bleu} to evaluate the results. For translation speed, we average the inference latency on the valid set of WMT14 EN$\rightarrow$DE three times and compare it to that of the AT model.

\subsubsection{Baselines}
Since our approach of EECR is a universal strategy that shrinks the training and inference discrepancy for the conditional masked language model, we apply it to two classical mask-predict-based models, CMLM~\cite{ghazvininejad-etal-2019-mask} and CMLMC~\cite{huang2021improving}, denoted as CMLM-EECR and CMLMC-EECR.

For CMLM-EECR, we report the results under $K$ = 10, which is optimal.
For CMLMC, we also set $K$ to 10 during training, which is different from the original CMLMC that performs one round of refinement based on full masked sequences. 

\begin{table}[htbp]
% \centering
\begin{center}
\caption{Performance comparison on IWSLT14 DE$\rightarrow$EN raw dataset.}
\scalebox{0.80}{
\begin{tabular}{c|c|c} 
\hline
\multirow{1}{*}{Models} &\multirow{1}{*}{Iter.} & \multicolumn{1}{c}{IWSLT14 DE$\rightarrow$EN}  \\
\hline
  % Transformer & N& 34.74   \\
  NAT-FT & 1 & 24.21 \\
  NAT-DCRF & 1 &29.99 \\
  GLAT & 1 & 32.49   \\ 
  \hline
  NAT-IR & 10 & 23.94 \\
  % CMLM & 4 & 30.42   \\ 
  CMLM & 10 & 32.10   \\ 
   % MvCR-NAT & 4 &30.58 \\
 MvCR-NAT & 10 & 32.55   \\ 
 \hline
 % CMLM-EECR & 4 & \textbf{30.78}\\
 CMLM-EECR & 10 & \textbf{32.82}   \\ 
\hline
\end{tabular}
}
\label{tab:iwslt}
\end{center}
\end{table}

To compare with previous methods, we select the AT Transformer as well as a series of NAT models as the baseline models. The NAT models include the fully NAT models as well as the iterative refinement NAT models.

\begin{table*}[thb]
\centering
\caption{Performance comparison between CMLM and our CMLM-EECR with/without Knowledge Distillation (KD) under inference iteration of 10.}
\scalebox{0.80}{
\begin{tabular}{l|c|cc|cc} 
\hline
\multirow{2}{*}{Model}& \multirow{2}{*}{Type}  & \multicolumn{2}{c|}{WMT'14} & \multicolumn{2}{c}{WMT'16}   \\
&    & EN$\rightarrow$DE  & DE$\rightarrow$EN  & EN$\rightarrow$RO & RO$\rightarrow$EN                       \\ 
\hline
\multirow{2}{*}{CMLM}   & Raw & 24.55 & 29.22 &  32.52 & 32.67     \\ 
   & KD & 26.98  & 30.84  & 33.15  & 33.41     \\ 
\hline
\multirow{2}{*}{CMLM-EECR}   & Raw &  25.33 &30.37& 33.27& 33.45   \\ 
   &KD & 27.95 &31.59 & 33.84 &34.21  \\ 
\hline

\end{tabular}
}
\label{tab:KD}
\end{table*}

\subsection{Main Results}

Table~\ref{tab:main} shows the main results of our model on the four distilled datasets and Table~\ref{tab:iwslt} displays the result on IWSLT14 DE$\rightarrow$EN raw dataset.

\subsubsection{Generality of our methodology}
Our model significantly improves the translation performance compared to the base models of CMLM and CMLMC. 
Specifically, our CMLM-EECR and CMLMC-EECR yield an average of 0.68 and 0.40 BLEU score improvements on four datasets compared to CMLM and CMLMC, respectively. 
This result indicates that our model does achieve performance improvement by narrowing down the difference in data distribution between training and inference. 

\subsubsection{Comparison with the SOTA models}

Our CMLMC-EECR attains comparable results with the AT Transformer and those strong competitors of fully NAT models. 
Specifically, our model CMLMC-EECR slightly outperforms the AT baseline on WMT16 EN$\rightarrow$RO and further closes the gap with the AT baseline on the remaining three datasets. 
Besides, our CMLMC-EECR excels over the fully NAT models, including GLAT-VAE and GLAT-DSLP in terms of translation quality. 

Our model reaches competitive performance among the iterative NAT models. Impressively, our best variant, CMLMC-EECR, gets SOTA performance on WMT14 EN$\rightarrow$DE and WMT16 EN$\rightarrow$RO. 
In addition, compared to some previous methods for narrowing the training/inference mismatch, our method is superior in performance. 
Our CMLM-EECR outperforms CMLM-SAMRT on WMT14 EN$\leftrightarrow$DE, and it surpasses MvCR-NAT on WMT14 EN$\rightarrow$DE and WMT16 EN$\leftrightarrow$RO, which further corroborates the effectiveness of our approach.

% This may be due to the fact that we perform consistency regularization over the probability distribution of the masked tokens under erroneous sequences as well as under ground truth sequences. 

% which introduces prediction errors into the training process as well. In addition,

\subsubsection{Performance on IWSLT14 DE-EN}
\label{iwslt}
We also compare the performance of CMLM-EECR with the existing models on the IWSLT14 DE$\rightarrow$EN raw dataset.
As seen in Table~\ref{tab:iwslt}, our CMLM-EECR attains 0.72 BLEU score increase compared to CMLM baseline and achieves considerable improvement over other previous competitors on IWSLT14 DE$\rightarrow$EN raw dataset.

\subsection{Ablation Study and Analysis}

\subsubsection{Loss weights for Consistency Learning}
Table~\ref{tab:Loss} presents the effect of consistency regularization intensity on translation quality. 
We set different weights for consistency regularization on WMT14 EN$\rightarrow$DE and WMT16 EN$\rightarrow$RO. 
The results reveal that our model obtains the best performance when the weight factor is 0.4. 
When the weight factor is too small ($e.g.$, 0.2), the model gets poor translation quality, which may be due to the insufficiency of consistency regularization. 
When the weight factor is too large ($e.g.$, 0.8), the consistency learning, as an auxiliary task, could affect the convergence of the main translation task.

\begin{table}[htb]
% \centering
\begin{center}
\caption{Loss weights for consistency learning of CMLM-EECR.}
\scalebox{0.80}{
\begin{tabular}{c|c|c} 
\hline
\multirow{1}{*}{Weight}  & \multicolumn{1}{c|}{WMT'14 EN$\rightarrow$DE} & \multicolumn{1}{c}{WMT'16 EN$\rightarrow$RO}  \\
\hline
 0.0 & 27.53 & 33.46   \\ 
 0.2 & 27.62 & 33.69   \\ 
 0.4 & \textbf{27.95} & \textbf{33.84} \\ 
 0.6 &27.76 & 33.62      \\ 
 0.8 &27.22 & 33.16     \\ 
\hline
\end{tabular}
}
\label{tab:Loss}
\end{center}
\end{table}

\subsubsection{The Effect of Knowledge Distillation}

In Table~\ref{tab:KD} we evaluate the model under raw corpus without knowledge distillation, and we find CMLM-EECR outperforms CMLM-EECR on raw dataset. 
We notice that in WMT16 EN$\leftrightarrow$RO, the performance of CMLM-EECR in the raw dataset even slightly outperforms that of CMLM in the KD dataset.

\begin{table}[htbp]
% \centering
\begin{center}
\caption{The effect of Consistency Regularization Terms.}
\scalebox{0.80}{
\begin{tabular}{l|c|c} 
\hline
\multirow{1}{*}{CR Term}  & \multicolumn{1}{c|}{WMT'14 EN$\rightarrow$DE} & \multicolumn{1}{c}{WMT'16 EN$\rightarrow$RO}  \\
\hline
$\mathcal{L}_{nll}$ & 27.53 & 33.46   \\ 
$\mathcal{L}_{nll}$+$\mathcal{L}_{kld}^{1}$ & 27.71 & 33.61   \\ 
$\mathcal{L}_{nll}$+$\mathcal{L}_{kld}^{2}$ & 27.78 & 33.56   \\ 
$\mathcal{L}_{nll}$+$\mathcal{L}_{kld}^{1}$+$\mathcal{L}_{kld}^{2}$ & \textbf{27.95} & \textbf{33.84}   \\ 
\hline
\end{tabular}
}
\label{tab:CR}
\end{center}
\end{table}

\begin{table}[htbp]
% \centering
\begin{center}
\caption{The effect of multi-iteration on IWSLT14 DE$\rightarrow$EN.}
\scalebox{0.80}{
\begin{tabular}{c|c|c|c} 
\hline
\diagbox{Infer Iter.}{Train Iter.}& 1 & 4 & 10  \\
\hline
 1 & \textbf{23.44} & 23.30 & 23.19 \\ 
 4 & 30.88&    \textbf{31.02} &  30.92 \\ 
10 & 32.53& 32.69  &     \textbf{32.82}\\ 
\hline
\end{tabular}
}
\label{tab:multi-iter}
\end{center}
\end{table}

\begin{table*}[thb]
\centering
\caption{Performance comparison between CMLM and our CMLM-EECR under different inference iterations.}
\scalebox{0.80}{
\begin{tabular}{l|c|cc|cc|c} 
\hline
\multirow{2}{*}{Model}& \multirow{2}{*}{Iter.}  & \multicolumn{2}{c|}{WMT'14} & \multicolumn{2}{c|}{WMT'16}  & \multirow{2}{*}{Speedup} \\
&    & EN$\rightarrow$DE  & DE$\rightarrow$EN  & EN$\rightarrow$RO & RO$\rightarrow$EN    &                     \\ 
\hline
   & 1 & 18.48 & 22.35 & 27.46 & 28.03 & 15.3$\times$     \\ 
 CMLM & 4 & 26.72 & 30.56 & 32.68 &33.29  & 9.79$\times$    \\ 
   & 10 & 27.25 & 31.08 & 33.15 & 33.41 & 3.77$\times$     \\ 
\hline
          & 1 & 19.23 & 23.53 & 29.13 & 30.68 & 15.3$\times$   \\ 
CMLM-EECR & 4 & 27.33 & 31.12 & 33.28 & 33.73 & 9.79$\times$      \\ 
           &10& 27.95 & 31.59 & 33.84 & 34.21 & 3.77$\times$   \\ 
\hline

\end{tabular}
}
\label{tab:iteration}
\end{table*}

\subsubsection{Consistency regularization Terms}
In our proposed methods, we construct two consistency regularization terms, including the constraint for the distribution of masked tokens under two different mixed sequences ($\mathcal{L}_{kld}^{1}$ in Eq.~\ref{eqn:kld1}), and that of the mixed sequences and the ground truth ($\mathcal{L}_{kld}^{2}$ in Eq.~\ref{eqn:kld2}). 
In Table~\ref{tab:CR} we explore the effect of these two consistency regularization terms on the model performance. 
$\mathcal{L}_{nll}$ denotes the sum of NLL losses in Eq.~\ref{eqn:nll1},~\ref{eqn:nll2} and~\ref{eqn:nll3}.
The experimental results suggest that using only one regularization term could improve the performance, and the best translation quality is achieved when combining the two consistency regularity terms.

\subsubsection{Multi-iteration in Mixed Sequence Construction}
In the construction process of mixed sequence, we introduce multiple rounds of refinement in sequence prediction and we further explore the effect of it in Table~\ref{tab:multi-iter}. 
For a given inference iteration, it achieves the highest translation quality when it equals the training iteration $K$. 
This may be due to the fact that the exposure bias between training and inference is minimized in this case.
We also discover that the highest BLEU score is obtained when the training and inference iteration numbers are both 10. 
We gain a similar trend on WMT14 EN$\leftrightarrow$DE and WMT16 EN$\leftrightarrow$RO, and part of the results are showcased in Appendix~\ref{app:multi-iter}.

\subsubsection{The Effect of Repetition}
Table~\ref{tab:Repetition} exhibits the word repetition rate of our model CMLM-EECR and the base model CMLM under different inference iterations. 
The results show that the word repetition rate of our model is significantly reduced compared with the base model. 
This suggests that our model could better capture the dependencies within the target sentence, thus reducing the modality in the target sentence and decreasing the word repetition rate. 

\begin{table}[htbp]
% \centering
\begin{center}
\caption{Repetition rate of CMLM and CMLM-EECR.}
\scalebox{0.70}{
\begin{tabular}{l|c|cc|cc} 
\hline
\multirow{2}{*}{Model}& \multirow{2}{*}{Iter.}  & \multicolumn{2}{c|}{WMT'14 EN$\rightarrow$DE} & \multicolumn{2}{c}{WMT'16 EN$\rightarrow$RO}  \\
\cline{3-6}
&    & BLEU  & Rep  &  BLEU & Rep                           \\ 
\hline
       & 1 & 18.48 & 17.84\% & 27.46 & 10.82\%     \\ 
 CMLM  & 3 & 25.36 & 1.85\% & 32.30 & 1.39\%     \\ 
       & 5 & 26.99 & 0.83\% & 32.79 &0.65\%    \\  
\hline
          & 1 & 19.23 & 11.61\% & 29.13 & 7.68\%   \\ 
CMLM-EECR  & 3 & 27.01 & 0.87\% & 32.91 & 0.65\%   \\ 
           & 5 & 27.42 & 0.66\% & 33.35 & 0.41\% \\
        
\hline

\end{tabular}
}
\label{tab:Repetition}
\end{center}
\end{table}

\subsubsection{Different Inference Iteration}
In Table~\ref{tab:iteration}, we compare the translation performance of the CMLM baseline with our CMLM-EECR under different inference iteration rounds. 
There are two indicators, BLEU and speedup, for the translation performance. 
Following the previous work~\cite{ghazvininejad-etal-2019-mask}, we set the number of iteration to 1, 4 and 10. 
As we can see from Table~\ref{tab:iteration}, as the iteration increases, the BLEU score of both models increases. 
We also notice that our EECR (iter=4) gets very close to CMLM (iter=10) on WMT14 DE$\rightarrow$EN and even exceeds the results of the CMLM (iter=10) on the remaining three datasets.

\begin{figure}
\includegraphics[width=0.45\textwidth]{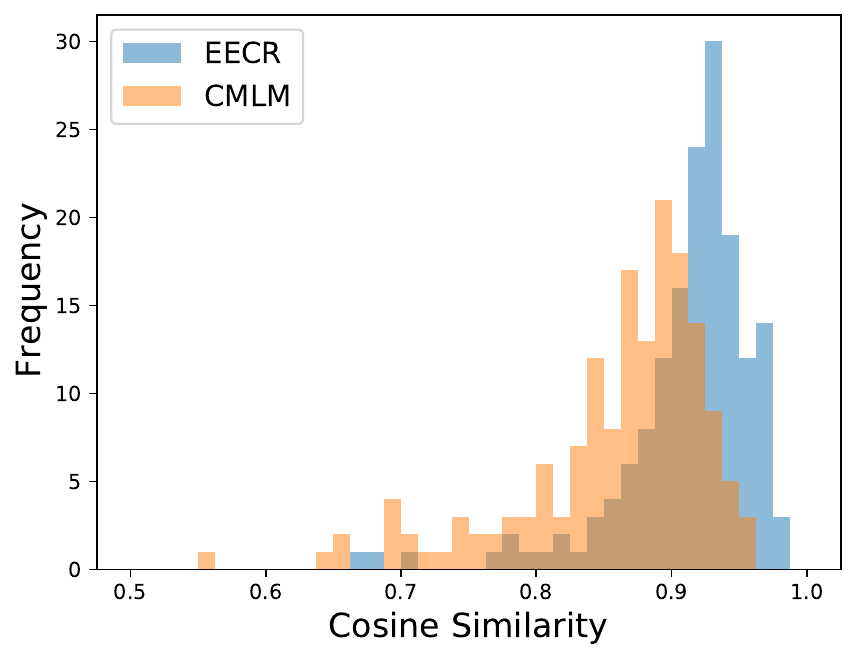}
\caption{The cosine similarity of masked token representations under different observing scenarios of CMLM-EECR and CMLM.}
\label{fig:simi}
\end{figure}

\subsubsection{The Similarity of Prediction}
\label{similarity}
We verify the effectiveness of our EECR model in reducing the exposure bias. 
We take the mixed sequence as well as the ground truth sequence with the same masked pattern as the model input and compute the cosine similarity of the output distribution probabilities of the masked tokens on CMLM-EECR and CMLM model.
The histogram of the probability distribution is presented in Figure~\ref{fig:simi}. 
We observe that EECR mechanism improves the similarity of the probability distributions of tokens under different input scenarios, which implies that CMLM-EECR has more consistent outputs, and suggests that EECR effectively reduces the exposure bias. 
We present more experimental results and analyses in Appendix~\ref{app:curve}, ~\ref{app:length} and~\ref{app:case}.

% The method constructs training examples containing errors and optimizes the model under imperfect conditions. 
% In addition, we adopt consistency regularization to make the probability distributions of the masked tokens in different observing conditions as consistent as possible,

\section{Conclusion}
In this paper, we propose a training strategy, EECR, based on error exposure and consistency learning to mitigate the training/inference mismatch problem of CMLM models. 
Our method is simple and widely applicable. 
In terms of error exposure, we construct mixed sequences containing potential errors as training samples by introducing multi-round refinement during training. 
Further, we introduce consistency regularization for the output distribution of the masked tokens under different observing situations. 
Experiments on widely-used datasets show that our method improves the translation quality compared to the baselines, regarding BLEU score and repetition rate. 
Meanwhile, our method could minimize the difference in the distribution of the model under different exposure scenarios. 
Further, our best variant close the gap with the performance of the AT model.

\section{Limitations}
Although our method of error exposure and consistency regularization mainly concentrates on improving the training of NAT and does not effect inference, it could increase the consumption of computing resources during training. 
In training, we perform multiple rounds of refinement to construct the predictions which extends the training time. 
In addition, the model is optimized based on the mixed sequences and ground truth, which also results in extra computational resource consumption. 
The comparison of training time on different models is shown in the Appendix~\ref{app:training time}.

\bibliography{custom}

\appendix
\section{Background}
\subsection{Non-autoregressive Machine Translation}
Autoregressive neural machine translation models generate translations in a left-to-right manner. Specifically, given a source sentence $X$, the AT models factorize the probability of the target sentence $Y$ with conditional dependency by:
\begin{equation}
P\left( Y|X \right) =\prod \limits_{t=1}^n P\left( y_{t}|y_{<t}, X \right),
\end{equation}
where $y_{<t}$ represents the sequence generated before time-step $t$ and where $n$ refers to the output length. The sequential decoding pattern of the AT model leads to long inference latency and inefficient usage of parallel hardware.

On the other hand, the NAT model, decodes in parallel based on the conditional independent assumption and improves translation efficiency greatly~\cite{gu2018non}. Unlike the AT model, which utilizes a special token $\texttt{[EOS]}$ as a decoding terminator, most NAT model determines the output length before decoding. The probability of NAT can be written as: 
\begin{equation}
P\left( Y|X \right) =P\left( n| X \right) \prod \limits_{t=1}^n P\left( y_{t}| X \right),
\end{equation}

Since the NAT model breaks the conditional dependency, it cannot model the dependency within the target sentences well, and there is room for translation quality improvement.

\subsection{Conditional Masked Language Model}
The CMLM model is an effective iterative-refinement-based NAT~\cite{ghazvininejad-etal-2019-mask}, which improves the translation quality by masking and re-predicting the low-confidence tokens in multiple rounds of iterations. 
In each iteration, CMLM predicts based on the output of the previous iteration, and captures the interdependency of the target sentence outperforming other NAT models~\cite{ghazvininejad-etal-2019-mask,kasai2020non}. 
The prediction probability of CMLM is:
\begin{equation}
P\left( Y|X \right) = -\prod \limits_{t=1}^{\lvert Y_{mask} \rvert}{ P\left(y|X,Y_{obs}\right)}
\end{equation}
where $Y_{obs}$ is the observed tokens and $Y_{mask}$ is the masked tokens. 
Please note that the observed tokens come from ground truth and model prediction during training and inference, respectively.

\section{Details}
\label{app:Details}
For WMT14 EN$\leftrightarrow$DE and WMT16 EN$\leftrightarrow$RO, the model structure is based on \texttt{transformer}~\cite{vaswani2017attention}, where both encoder and decoder are stacked by 6 transformer layers and the model dimension and hidden dimension are 512 and 2048, respectively. 
For IWSLT14 DE$\rightarrow$EN, we configure the model based on the traditional \texttt{transformer\_iwslt\_de\_en} setting, where both encoder and decoder are stacked by 6 transformer layers and the model dimension and hidden dimension are 256 and 1024, respectively. 
For training hyper-parameters, we set the dropout rate to 0.3 for WMT16 EN$\leftrightarrow$RO and IWSLT14 DE$\rightarrow$EN, 0.2 for WMT14 EN$\leftrightarrow$DE.
We configure the weight decay to 0.01, and set the label smoothing to 0.1. 
We utilize the Adam~\cite{kingma:adam} optimizer with $\beta$ = (0.9, 0.98), $\epsilon = 10^{-6}$. 
We set the learning rate to $5\cdot 10^{-4}$ for WMT16 EN$\leftrightarrow$RO and IWSLT14 DE$\rightarrow$EN, $7\cdot 10^{-4}$ for WMT14 EN$\leftrightarrow$DE. 
The learning rate grows in the initial 10k steps and decays in subsequent updates with an inverse square-root schedule. 
We set the tokens per training batch to 128k for WMT14 EN$\leftrightarrow$DE, 32k for WMT16 EN$\leftrightarrow$RO, 8k for IWSLT14 DE$\rightarrow$EN. 
We train our models on the NVIDIA GeForce RTX 3090 GPU and set the maximum training update to 150k. 
During inference, we adopt noisy parallel decoding (NPD)~\cite{gu2018non} strategy to generate 5 candidates of different lengths and then select the one with the best quality as the final output. 
Following the previous works~\cite{xie2022mvsr}, we average the last 10 checkpoints to produce the final translation.

\section{Multi-iteration in Mixed Sequence Construction}
\label{app:multi-iter}

We present the results on the WMT14 DE$\rightarrow$EN as well as the WMT16 RO$\rightarrow$EN in Table~\ref{tab:multi-iter2} and~\ref{tab:multi-iter3}.
The highest BLEU score is obtained when the training and inference iteration number are both 10.

\begin{table}[htbp]
% \centering
\begin{center}
\caption{The effect of multi-iteration on WMT14 DE$\rightarrow$EN.}
\scalebox{0.80}{
\begin{tabular}{c|c|c|c} 
\hline
\diagbox{Infer Iter.}{Train Iter.}& 1 & 4 & 10  \\
\hline
 1 & 24.52  & 24.40 & \textbf{24.73}\\ 
 4 & 30.55&    \textbf{30.72} & 30.68 \\ 
10 & 31.42&   31.44&  \textbf{31.59}\\ 
\hline
\end{tabular}
}
\label{tab:multi-iter2}
\end{center}
\end{table}

\begin{table}[htbp]
% \centering
\begin{center}
\caption{The effect of multi-iteration on WMT16 RO$\rightarrow$EN.}
\scalebox{0.80}{
\begin{tabular}{c|c|c|c} 
\hline
\diagbox{Infer Iter.}{Train Iter.}& 1 & 4 & 10  \\
\hline
 1 & \textbf{30.68} & 30.25 & 30.48\\ 
 4 & 33.56 &  \textbf{33.73} &  33.67 \\ 
10 &  33.93 &  34.08   &    \textbf{34.21}\\ 
\hline
\end{tabular}
}
\label{tab:multi-iter3}
\end{center}
\end{table}

\section{Baseline Comparison}
\label{app:base}

\begin{table*}[thb]
\centering
\caption{Comparison between the results of our re-implementation and that reported in the original paper. $^*$ represents the results under our implementation.}
\scalebox{0.80}{
\begin{tabular}{l|c|cc|cc} 
\hline
\multirow{2}{*}{Model}& \multirow{2}{*}{Iter.}  & \multicolumn{2}{c|}{WMT'14} & \multicolumn{2}{c}{WMT'16}  \\
&    & EN$\rightarrow$DE  & DE$\rightarrow$EN  & EN$\rightarrow$RO & RO$\rightarrow$EN                         \\ 
\hline
 CMLM~\cite{ghazvininejad-etal-2019-mask}  & 10 & 27.03 & 30.53 & 33.08 & 33.31   \\ 
 CMLM$^*$  & 10 & 27.25 & 31.08 & 33.15 & 33.41    \\ 
\hline
CMLMC~\cite{huang2021improving} & 10 & 28.37 & 31.41 & 34.57 & 34.13 \\
CMLMC$^*$  & 10 & 27.38 & 31.16 & 34.08 & 34.15  \\ 
\hline

\end{tabular}
}
\label{tab:basecompare}
\end{table*}
Table~\ref{tab:basecompare} displays the performance comparison of our reproduced baselines, CMLM and CMLMC, with the results from the original paper.

\section{The Training Curve of CMLM-EECR}
\label{app:curve}
\begin{figure}
\includegraphics[width=0.45\textwidth]{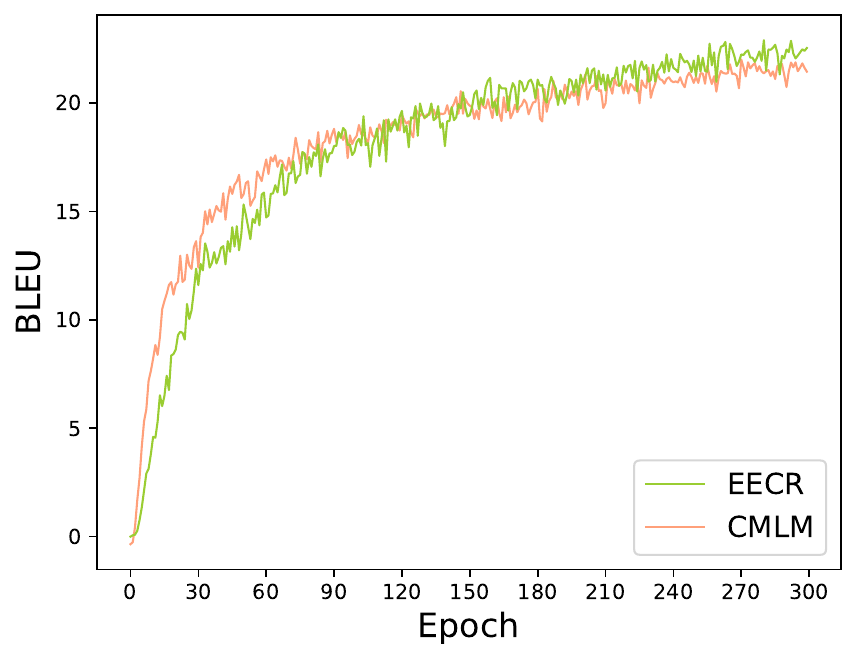}
\caption{The training curves of CMLM and CMLM-EECR in IWSLT14 DE$\rightarrow$EN valid set. The inference iteration number is set to 1.}
\label{fig:training curve}
\end{figure}

Figure~\ref{fig:training curve} reveals the increasing trend of BLEU score with training epochs in the IWSLT14 DE$\rightarrow$EN valid set. To make a fair comparison with CMLM-EECR, we triple the number of sentences in the forward pass for the baseline model. 
From Figure~\ref{fig:training curve}, we observe that the BLEU growth of the EECR model is slower than that of the baseline model before epoch 150, which could be restricted by the consistency regularization method. After 150 epochs, the performance of EECR exceeds that of the baseline model, which further proves the effectiveness of our model.

\section{The Effect of Sentence Length}
\label{app:length}

\begin{figure}
\includegraphics[width=0.45\textwidth]{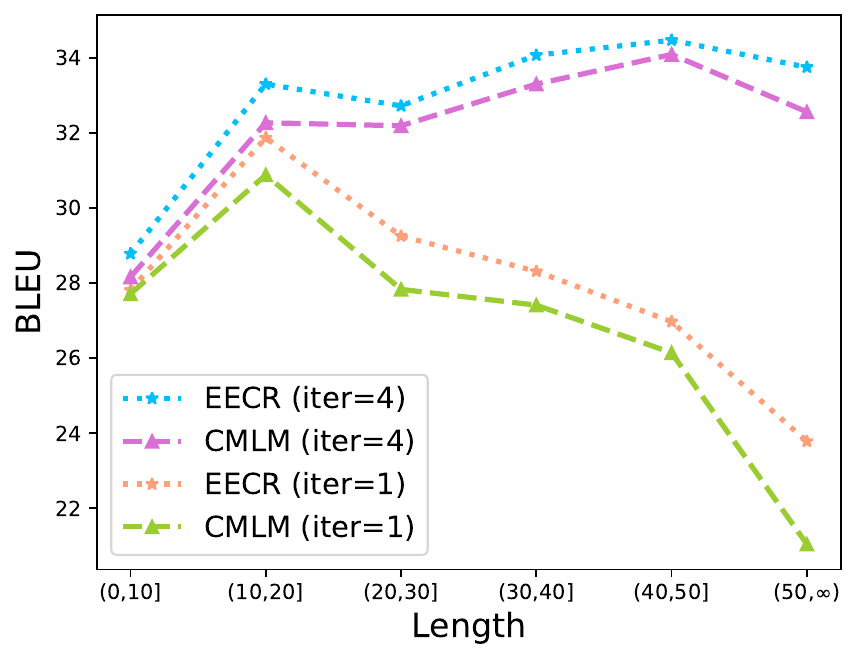}
\caption{Translation quality on WMT16 EN$\rightarrow$RO
test set over the sentence groups of different lengths.}
\label{fig:length}
\end{figure}

We further study the translation effect of the EECR method on different sentence lengths. 
We divide the test set of WMT16 EN$\rightarrow$RO into 6 buckets according to the length of the source sentences and compare the translation result. 
Figure~\ref{fig:length} reveals that our CMLM-EECR exceeds the CMLM baseline on all length groups at decoding iteration of 1 and 4, which verifies the effectiveness of our model.

\begin{table*}[thb]
\centering
\caption{Two translation examples of CMLM and our CMLM-EECR under different iterations. We bold the repeated words ($e.g.$, paper) and use underscores ($i.e.$, \_) to connect the sub-words.}
\scalebox{0.90}{
\begin{tabularx}{\hsize}{l|c|X}
\hline
\multicolumn{2}{l|}{Source} & 
Es ist ja nicht nur so , dass das Papierzeugs die Landschaft verschandelt .
 \\ 
 \hline
 \multicolumn{2}{l|}{Target} & After all , waste paper does more than spoil the landscape . \\
\hline
    &  1  & It is not only the case that the pap\_paper is sc\_ing the landscape . \\ 
 CMLM & 4  & It is not only the case that the \textbf{paper} \textbf{paper} is sc\_ing the landscape .  \\ 
    & 10  &  It is not only the case that the \textbf{paper} \textbf{paper} is wast\_ing the landscape .  \\ 
\hline
  & 1 & It is not only the case that \textbf{paper paper} bur\_utes the landscape . \\ 
CMLM-EECR & 4 & It is not only the case that the paper bur\_utes the landscape .    \\ 
  &10 & It is not only the case that the paper poll\_utes the landscape .  \\ 
 \hline
\multicolumn{2}{l|}{Source} & Selbst wer sich den Altersruhesitz in der Toskana oder der Bretagne leisten könne , sollte sich immer überlegen , dass er auch krank werden könne . \\ 
 \hline
 \multicolumn{2}{l|}{Target} & Even if you can afford to spend your retirement in Tuscany or Brittany , you should always take the fact that you may become ill into consideration . \\
\hline
  & 1 & Even those who can afford the \textbf{retirement retirement} in Tuscany or Britt\_Britt\_any should always \textbf{that that} he can become ill .  \\ 
 CMLM & 4  & Even if who can afford the \textbf{retirement retirement} in Tuscany or Britt\_any , \textbf{always always} \textbf{that that} he can become ill . \\ 
 & 10  &   Even those who can afford the \textbf{retirement retirement} in Tuscany or Britt\_any , should always consider that they can become ill . \\ 
\hline
  & 1 & Even those who can afford the \textbf{retirement retirement} in Tuscany or Britt\_any should always that they can fall ill \\ 
CMLM-EECR & 4  &  Even those who can afford the retirement in \textbf{Tuscany Tuscany} or Britt\_any should always that they can fall ill .  \\ 
 &10 &  Even those who can afford old retirement in Tuscany or Britt\_any should always consider that they can fall ill .  \\ 
 \hline
\end{tabularx}
}
\label{tab:example}
\end{table*}

\section{Case Study}
\label{app:case}

To demonstrate the translation quality of our model more intuitively, we compare the translation results of CMLM-EECR and the CMLM baseline model under different iterations in WMT14 DE$\rightarrow$EN test set in Table~\ref{tab:example}. 
As the number of decoding iteration increases, the repetitions in the output of CMLM and CMLM-EECR both reduce. 
However, in CMLM, there are still repetitive words in the final result even after 10 rounds of refinement, while the final translation of our CMLM-EECR contains fewer duplicated words, demonstrating our superiority in translation quality.

\begin{table}[b]
\centering
\begin{center}
\caption{Training time for CMLM baseline and CMLM-EECR under different training iterations ($K$).}
\scalebox{0.80}{
\begin{tabular}{l|c|c} 
\hline
\multirow{1}{*}{Model}  & \multicolumn{1}{c|}{WMT'14 EN$\leftrightarrow$DE} & \multicolumn{1}{c}{WMT'16 EN$\leftrightarrow$RO}  \\
\hline
 CMLM           &  17   & 8  \\ 
 EECR ($K=1$) & 32 & 9 \\ 
 EECR ($K=4$) & 36 & 10  \\ 
 EECR ($K=10$)  & 48 & 13  \\ 
\hline
\end{tabular}
}
\label{tab:training time}
\end{center}
\end{table}

\section{Training Time}
\label{app:training time}

We report the training time for CMLM-EECR and CMLM baseline on a machine with 4 NVIDIA GeForce RTX 3090 GPUs in Table~\ref{tab:training time}. 
% In IWSLT14 DE$\rightarrow$EN, we set 4096 tokens/batch on a single GPU and update frequency is set to 2. 
In WMT16 EN$\leftrightarrow$RO, we set 4096 tokens/batch on 2 GPUs and set update frequency to 4. 
On WMT14 EN$\leftrightarrow$DE, we set 4096 tokens/batch on 4 GPUs and set update frequency to 8. 
We can see that EECR leads to longer training time compared to the baseline model. 
Specifically, the introduction of mixed sequence extends training compared to that only on the ground truth as a training sample (EECR $v.s.$ CMLM). Additionally, a larger training iteration also results in a longer training time ($K=4, 10$ $v.s.$ $K$=1).

\end{document}